\documentclass[10pt,twocolumn,letterpaper]{article}
\usepackage{multirow}

\usepackage[pagenumbers]{cvpr} 

\usepackage[dvipsnames]{xcolor}

\definecolor{cvprblue}{rgb}{0.21,0.49,0.74}
\usepackage[pagebackref,breaklinks,colorlinks,citecolor=cvprblue]{hyperref}

\def\paperID{19} 
\def\confName{CVPR}
\def\confYear{2024}

\title{Dynamic Distinction Learning:\\Adaptive Pseudo Anomalies for Video Anomaly Detection}

\author{Demetris Lappas\\
{\tt\small k1838447@kingston.ac.uk}
\and
Vasileios Argyriou\\
{\tt\small vasileios.argyriou@kingston.ac.uk}
\and
Dimitrios Makris\\
{\tt\small d.makris@kingston.ac.uk}\\
Kingston University, London, UK,\\
School of Computer Science and Mathematics
}

\begin{document}
\maketitle

\begin{abstract}
We introduce Dynamic Distinction Learning (DDL) for Video Anomaly Detection, a novel video anomaly detection methodology that combines pseudo-anomalies, dynamic anomaly weighting, and a distinction loss function to improve detection accuracy. By training on pseudo-anomalies, our approach adapts to the variability of normal and anomalous behaviors without fixed anomaly thresholds. Our model showcases superior performance on the Ped2, Avenue and ShanghaiTech datasets, where individual models are tailored for each scene. These achievements highlight DDL's effectiveness in advancing anomaly detection, offering a scalable and adaptable solution for video surveillance challenges. Our work can be found on: \small\url{https://github.com/demetrislappas/DDL.git}
\end{abstract}

\section{Introduction}

Anomaly detection is pivotal in the field of video surveillance, where algorithms scan through endless hours of footage to identify activities or events that deviate from the norm—be it unauthorized intrusions, unusual behavior, or safety breaches. Its application in video analysis is indispensable across a multitude of sectors, underpinning security protocols, ensuring public safety, and enhancing operational efficiency. The capacity of video anomaly detection systems to flag deviations in real-time or in hindsight allows organizations to take quick, informed action to mitigate risks.

Nevertheless, the task of distinguishing the ordinary from the extraordinary in videos is exceptionally challenging. Video anomaly detection typically lives within the domain of unsupervised learning due to the inherent scarcity of labeled anomalies and the impracticality of cataloging the large array of possible anomalous events. The unpredictable nature of anomalies further adds to the complexity, making it difficult for models trained on `normal' behavior to generalize and identify outliers effectively. This difficulty is magnified by the context-sensitive definition of what constitutes an anomaly within video data, as it can vary significantly from one setting to another. In the absence of sufficient examples of anomalous behavior during training, systems often struggle to accurately discern anomalies when they do occur, resulting in a high number of false positives or missed detections.

Traditional approaches to this challenge have relied on neural network architectures like AutoEncoders and UNets \cite{Park2020,Saypadith2021,Deepak2021a,Chang2022,Chandrakala2022,Deepak2021,Feng2021,Massoli2020,Yuan2021,Vu2021,Liu2021,Roy2021,Szymanowicz2022,Kim2022,Le2022,Wang2021}. These models are trained to recreate `normality' by learning to compress and then reconstruct input data with minimal loss. The underlying premise is that, by becoming adept at reconstructing normality, these networks would inherently struggle when faced with anomalies, thus allowing for their detection. However, there lies a catch—these systems do not necessarily learn an explicit distinction between normal and anomalous samples, it is only \textit{hoped} that anomalies will pose a greater challenge for the reconstruction process. 

To address this predicament, various methodologies have introduced pseudo-anomalies during the training phase, offering models a taste of the `abnormal' to foster learning \cite{Zaheer2020,Astrid2022,Astrid2023,barbalau2023ssmtl++,Aich_2023_WACV,Liu_2023_CVPR}. These strategies, however, often overlook a critical aspect: the quantification of the `right level' of pseudo-anomaly. That is, how anomalous should pseudo-anomalies be to represent real anomalies? Too small, and the pseudo-anomalies bare too close a resemblance to normal data; too high, and the model may fail to recognize genuine, more subtle, anomalies.

In our work, the innovation lies not just in the incorporation of pseudo-anomalies, but in the strategic introduction of a dynamic anomaly weight $\sigma(\ell)$. This adaptability is crucial, allowing our model the flexibility to discover the optimal threshold of anomaly intensity for effective learning. Rather than being constrained to a predetermined, static level of pseudo-anomaly — which might risk the model's overfitting to artificial quirks — the dynamic nature of $\sigma(\ell)$ entrusts the model with the autonomy to fine-tune this threshold. By doing so, the model is trained to differentiate between normal and anomalous patterns without being anchored to any specific level of anomaly defined by the user.

Our work also introduces the Distinction Loss, which works in tandem with $\sigma(\ell)$, and is crafted to refine the model's discrimination capabilities. The Distinction Loss encourages the model to rebuild pseudo-anomalous frames to more closely resemble the normal state rather than the inputted anomalous one.

 In the forthcoming chapters, we delve into the core of our research on Dynamic Distinction Learning (DDL) for video anomaly detection. We first begin by providing a brief overview of related work in Chapter \ref{sec:rel_work}. Chapter \ref{sec:methodology} outlines the methodology, detailing the DDL framework and its components. Chapter \ref{sec:datasets} describes the datasets considered for evaluation, leading to Chapter \ref{sec:results}, which presents our findings through quantitative results. In our final chapter, Chapter \ref{sec:ablations}, we provide ablation studies, highlighting the improvements offered by our work.

\section{Related Work}
\label{sec:rel_work}

The challenge of anomaly detection in video data is exacerbated by the predominance of normal behavior within datasets, leading to an inherent bias towards non-anomalous examples. Unsupervised learning, particularly through the use of AutoEncoders (AEs), has emerged as a preferred solution. AEs leverage the discrepancy between input and reconstructed output to identify anomalies, operating under the principle that unfamiliar anomalous inputs will result in significant reconstruction errors \cite{Park2020,Saypadith2021,Deepak2021a,Chang2022,Chandrakala2022,Deepak2021,Feng2021,Massoli2020}. However, the challenge of accurately reconstructing normal samples to distinguish them from anomalies remains, with UNets and their skip connections offering a partial solution by improving reconstruction fidelity, albeit complicating the reliance on the latent space for anomaly detection \cite{Yuan2021,Vu2021,Liu2021,Roy2021,Szymanowicz2022,Kim2022,Le2022,Wang2021}.

Recent advancements have explored the temporal dimension of video anomaly detection, employing AEs and UNets to reconstruct sequences or predict subsequent frames, under the hypothesis that anomalies will disrupt the model's ability to accurately predict future frames based on a sequence of normal frames \cite{Kim2022,Le2022,Szymanowicz2022,Chang2022,Wang2021,Liu2021,Vu2021,Saypadith2021,Roy2021,Yuan2021,Feng2021,Liu2023}. The integration of Transformers and attention mechanisms aims to capture the temporal characteristics of video data more effectively, enabling AutoEncoders and UNets to identify anomalies by focusing on the relationships between frames \cite{Yuan2021,Feng2021,Le2023,Ullah2023}. Optical Flow has been utilized to enhance motion-related anomaly detection, providing a compact yet informative representation of temporal changes by capturing pixel motion between consecutive frames \cite{Chang2022,Saypadith2021,Vu2021,Cai2021,Wang2023}. 

To improve the performance of AEs and UNets, some studies have incorporated supervised learning techniques, Generative Adversarial Networks (GANs), and Object Detection to refine the distinction between normal and anomalous samples. GANs, in particular, create a generative-discriminative adversarial relationship that enhances the model's ability to reconstruct outputs indistinguishable from the original input \cite{Vu2021,Saypadith2021,Roy2021,Yuan2021,Feng2021,Li2023}. Object Detection focuses the anomaly detection process on significant frame objects, albeit limited by the detection model's scope and accuracy \cite{Vu2021,Saypadith2021,Roy2021,Yuan2021,Feng2021,Li2023,Roy2021,Chandrakala2022,Doshi2021,Sun2023}. Memory modules have also been proposed to prevent anomaly reconstruction by referencing normal samples, suggesting enhanced model complexity as a pathway to more effective anomaly detection \cite{Massoli2020,Park2020,Liu2021,Szymanowicz2022,Chandrakala2022,Gong2019,Sun2023,Wang2023}.

The imbalance between normal and anomalous samples in datasets has necessitated the development of innovative approaches that introduce pseudo-anomalies. These methods are designed to enhance the capability of reconstruction-based models to distinguish between normal and anomalous samples with greater precision. Techniques for generating pseudo-anomalies vary widely, some strategies involve the use of external datasets to inject anomalies into a dataset of normal samples. This can involve leveraging attention mechanisms to identify and transfer key features from third-party datasets to normal samples, thus creating pseudo-anomalies \cite{Aich_2023_WACV}, or introducing noise into the latent space of models using external data \cite{Liu_2023_CVPR}. Other methods consider a more creative approach, which utilize the previous state of the model to generate lower quality reconstructions which would be represented as anomalous samples \cite{Zaheer2020}. More traditional approaches attempt to invoke abnormality during training by directly providing the model with human defined anomalous behavior, such as reversing the sequence of input frames \cite{Astrid2022}. More recent pseudo anomalous methods attempt to attain superior results by injecting a suite of human defined anomalies, including the manipulation of video frames by reversing their sequence, skipping frames, adding noise, fusing frames, or incorporating random patches \cite{Astrid2023,barbalau2023ssmtl++}.
Anomalies, regardless of their specific nature (skipping frames, repeating frames, introducing extraneous shapes, etc), are perceived by convolutional layer kernels as unusual collections of vector representations—noise. Despite their efficacy, these methods rely on manual intervention to simulate anomalies, requiring a subjective determination of the degree of anomaly introduced—raising the question, ``What constitutes the appropriate level of noise to be considered anomalous?"

Against this backdrop, our research introduces a sophisticated approach that not only incorporates the concept of dynamic anomaly weighting but also presents a novel distinction loss function. This methodology aims to advance the anomaly detection domain by providing a more refined mechanism for distinguishing between normal and anomalous events, thereby segueing into the detailed explanation of our proposed methodology outlined in Section \ref{sec:methodology}.

\section{Methodology}
\label{sec:methodology}

\begin{figure*}[t!]
\begin{center}
\includegraphics[width=\textwidth]{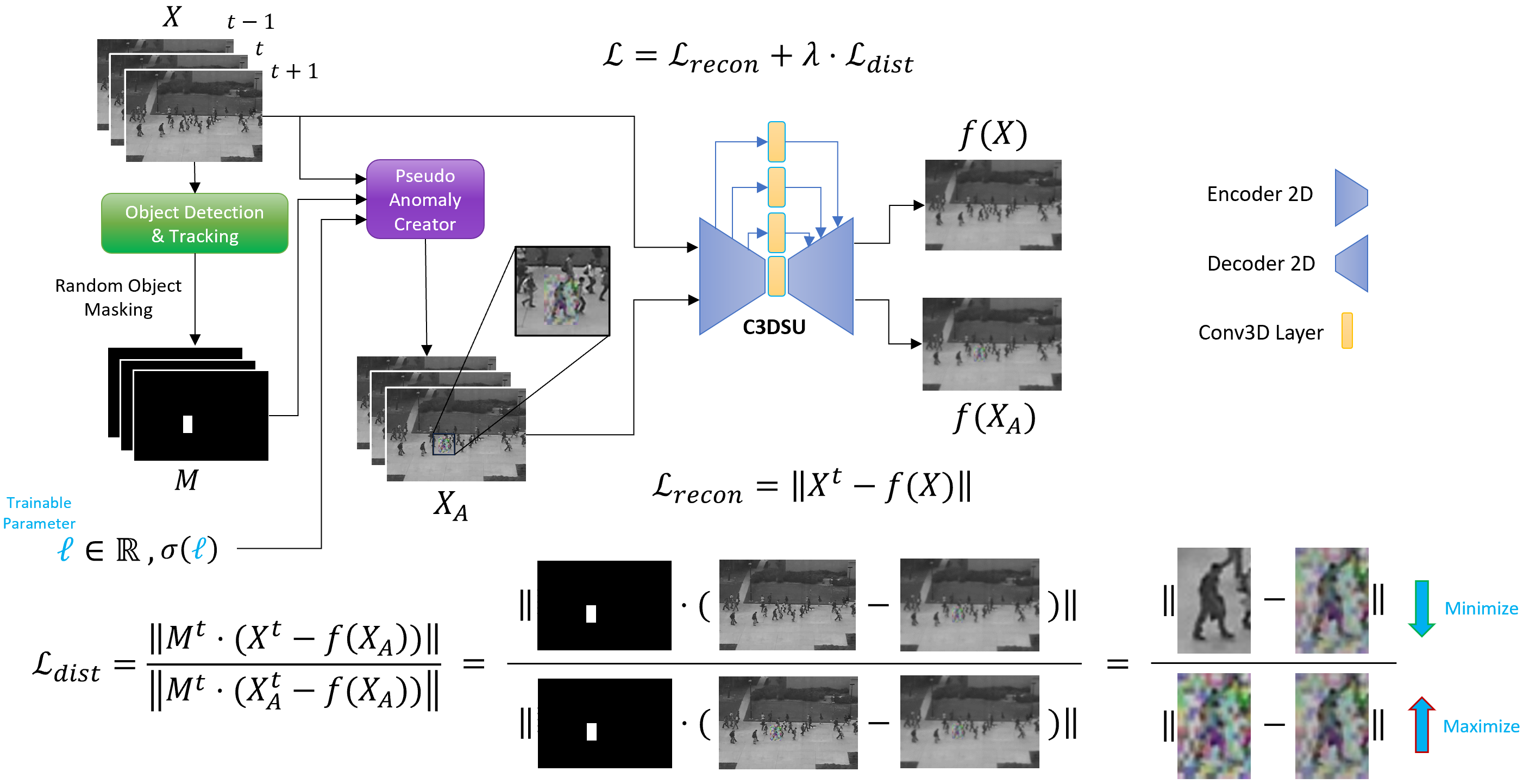}
\caption{The Dynamic Distinction Learning (DDL) Architecture: This diagram illustrates the DDL model's workflow, including object detection and tracking, random object masking, pseudo anomaly creation, our C3DSU model and the distinction loss calculation. The architecture depicts how the pseudo anomalies are created, then passed through the model along with their normal counter parts. The diagram also provides a visual depiction of the distinction loss calculation, showing how the model learns to minimize the numerator and maximize the denominator.}
\label{fig:ddl_architecture}
\end{center}
\end{figure*}

The Dynamic Distinction Learning (DDL) architecture is outlined in Figure \ref{fig:ddl_architecture}. Consider a sequence of normal video frames represented as a tensor $X \in \mathbb{R}^{c\times T \times H \times W}$, where $c$ is the number of channels, $T$ is the number of frames (which must be an odd number, as we will be reconstructing the middle frame), and $H$ and $W$ are the height and width of the frames respectively.  To simulate anomalies over the sequence, we pass the model through an Object Detection and Tracking model, followed by Random Object Masking - which selects a random tracked object across all frames and returns a sequence of binary masks $M \in \{ 0,1 \}^{c\times T \times H \times W}$ delineating the regions of the frames where the pseudo-anomaly will be present.
Alongside, we also introduce a noise tensor $A \in \mathbb{R}^{c\times T \times H \times W}$, which is uniformly random generated.

We also define a trainable parameter $\ell \in \mathbb{R}$, which
is passed through a sigmoid function, $\sigma(\ell) \in (0,1)$, to represent the anomaly weight. We chose a sigmoid function to bound the trainable parameter between the values of $0$ and $1$, so to portray a percentage of anomaly inflicted.  The sequence of normal frames $X$, the masks $M$, the noise tensor $A$, and the anomaly weight $\sigma(\ell)$ are passed into the Pseudo Anomaly Creator to fabricate pseudo anomalies $X_A$.

Both the sequence of normal frames $X$, and the pseudo anomalies $X_A$, are passed through a reconstruction model and calibrated using a linear combination of the Reconstruction Loss and Distinction Loss. The anomaly weight is heavily calibrated by the Distinction Loss, a loss function designed to converge the anomaly weight to represent the minimum anomaly capable of being detected. The adaptability of the anomaly weight $\sigma(\ell)$ allows the model to dynamically calibrate the degree of anomaly present in the training data, ensuring an effective balance between the recognition of normal patterns and the detection of deviations. This is critical for preventing the model from either becoming desensitized to subtle anomalies or overreacting to minor irregularities, thus maintaining a nuanced representation of what constitutes an anomaly throughout the training process.

\begin{figure*}[t!]
\begin{center}
\includegraphics[width=\textwidth]{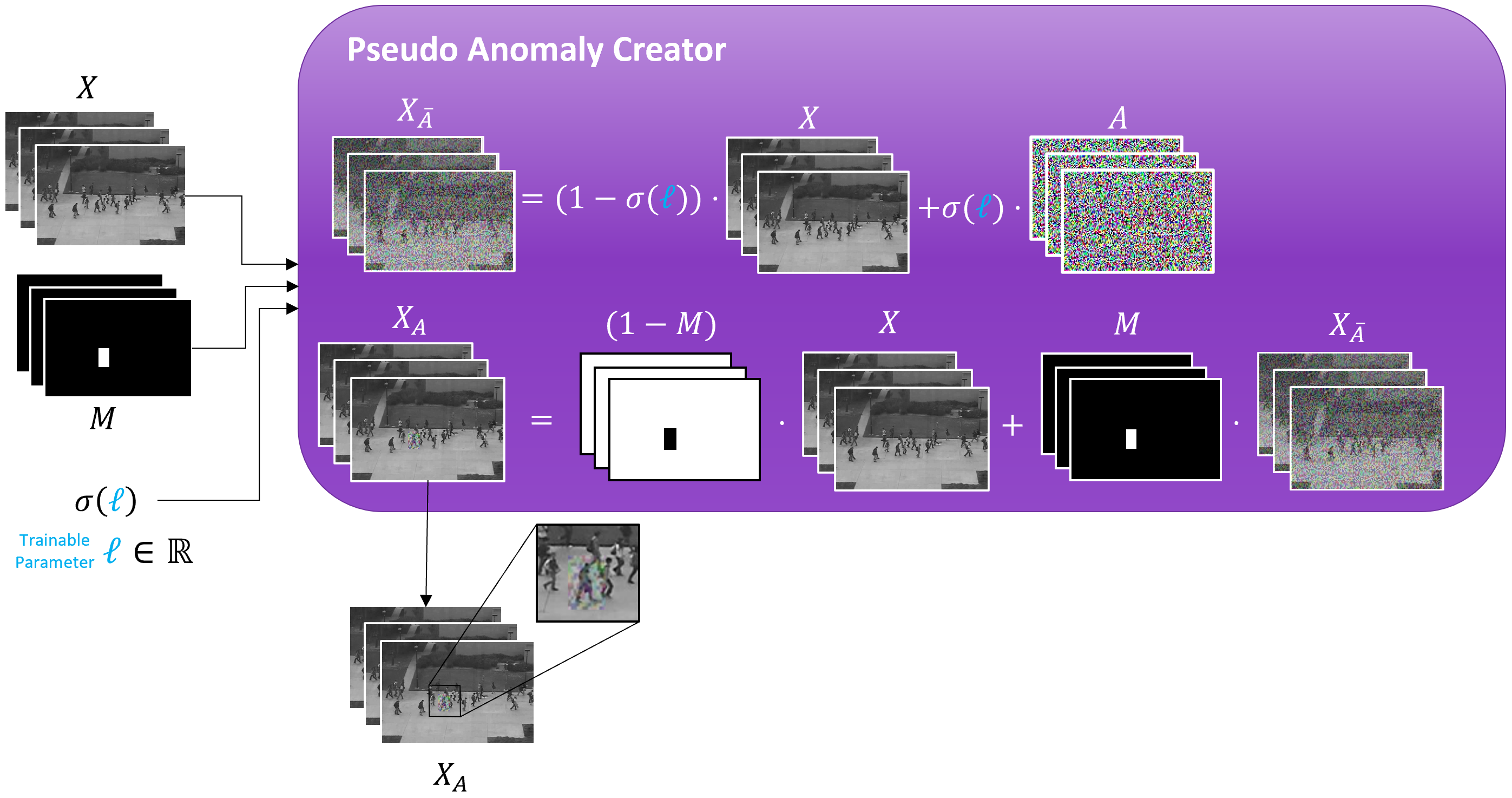}
\caption{Pseudo-Anomaly Creation Process: This figure demonstrates the step-by-step procedure for generating pseudo-anomalies within video frames. It begins by receiving the normal input frames, the masked frames, and a dynamically learned anomaly weight followed by the application of a noise tensor modulated by the anomaly weight.}
\label{fig:pseudo_anomaly_creator}
\end{center}
\end{figure*}

\subsection{Pseudo Anomaly Creator}

Our approach to fabricating pseudo-anomalies within video sequences begins with the application of object detection and tracking at each frame, then randomly selecting an object from the set of tracked objects for masking. We employ object tracking to consistently mask the same object across all frames within a temporal window, $T$. These masks, denoted as $M$, are crucial in defining the regions for anomaly simulation, ensuring the anomalies are contextually integrated around objects.

Following the identification and masking of objects, we proceed to the creation of pseudo-anomalies, via the Pseudo Anomaly Creator, a two-step noise integration process shown in Figure \ref{fig:pseudo_anomaly_creator}. Initially, we generate noise-infused frames, $X_{\bar{A}}$, by blending the original input frames, $X$, with a noise tensor, $A$, using the dynamically learned anomaly weight, $\sigma(\ell)$. This blend is achieved through a linear combination, ensuring the proportionate integration of noise and original content as per the following equation:

\begin{equation}
X_{\bar{A}} = (1-\sigma(\ell)) \cdot X + \sigma(\ell) \cdot A
\end{equation}

Here, the element-wise multiplication ($\cdot$) facilitates the precise control over the extent of noise addition, allowing for variable distortion levels that are directly influenced by the anomaly weight, which evolves during the training phase.

The subsequent phase involves the formulation of the pseudo-anomalous frames, $X_A$. These frames emerge from overlaying the noise-infused frames, $X_{\bar{A}}$, onto the original input frames, $X$, strictly within the boundaries defined by the object masks $M$. The mathematical representation of this process is captured by:

\begin{equation}
X_A = (1-M) \cdot X + M \cdot X_{\bar{A}}
\end{equation}

Through this method, we ensure that the noise, symbolizing potential anomalies, is selectively applied to the areas of interest - those being the detected objects within the frame. This approach not only maintains the contextual relevance of the introduced anomalies but also simulates a variety of anomalous patterns by leveraging the variability in noise composition; we elaborate on this in Section \ref{subsec:rationale_weighted_noise} within the Supplementary Material. By focusing on object regions, our method aims to create realistic and pertinent anomalies, enhancing the model's ability to detect and learn from these fabricated irregularities, which are designed to mimic a diverse spectrum of anomalous behaviors and appearances, including unseen shapes and uncommon motion blurs. 

\subsection{Reconstruction Model Definition}
\label{subsec:model_def}

We define a reconstruction model $f = \mathcal{E} \circ \mathcal{D}$, where $\mathcal{E}$ and $\mathcal{D}$ represent some encoder and decoder parts of a deep learning architecture, respectively. The choice of architecture is flexible and can include, but is not limited to, AutoEncoders, UNet structures, or other suitable convolutional neural networks designed for video reconstruction.

In practice we employ an adaptation of a 2D UNet model, tailored for the analysis of temporal data through the integration of 3D convolutional layers between skip connections. We call this architecture a \textit{Conv3DSkipUNet} (C3DSU or $f$ for the context of this work); more detail of our architecture can be found in Section \ref{C3DSU_Arch} within the Supplementary Material. The model receives an input such as $X$ and returns a reconstructed output $f(X) \in  \mathbb{R}^{c\times t \times H \times W}$, where $t$ represents the middle frame in $T$; that is, the model receives an odd sequence of frames as an input and returns the reconstructed middle frame.

\subsection{Loss Function}
We define a loss function, $\mathcal{L}$, which integrates the standard reconstruction loss with our novel distinction loss to fine-tune the model's sensitivity to anomalies.

\begin{equation}
\mathcal{L} = \mathcal{L}_
{recon} + \lambda \cdot \mathcal{L}_{dist}
\end{equation}

\noindent
where $\lambda$ is a hyperparameter that modulates the impact of the distinction loss relative to the reconstruction loss. This adjustment is crucial for ensuring that the model effectively balances learning to reconstruct normal frames while also distinguishing them from pseudo-anomalous frames.

\subsubsection{Reconstruction Loss}

The first function is the standard reconstruction loss:

\begin{equation}
    \mathcal{L}_{recon} = \lVert X^{t} - f(X) \rVert 
\end{equation}

\noindent
where $X^{t}$ is the middle frame of $X$. This loss function encourages the model to accurately reconstruct the normal input frame, thus learning the distribution of normal frames. 

\subsubsection{Distinction Loss}

The distinction loss is the second loss function in our model, designed to fine-tune the distinction between normal frames and their pseudo-anomalous counterparts. This differentiation is crucial for the model to recognize and identify anomalies effectively. The distinction loss function is articulated through the following mathematical formulations:

\begin{equation}
    P = \lVert M^{t} \cdot (X^{t} - f(X_A)) \rVert 
\end{equation}

\begin{equation}
    N = \lVert M^{t} \cdot (X^{t}_A - f(X_A)) \rVert 
\end{equation}

\begin{equation}
    \mathcal{L}_{dist} = \frac{P + \epsilon}{N + \epsilon}
\end{equation}

Here, $P$ serves to penalize the differences between the original normal frame $X^t$ and the model's reconstruction of the pseudo-anomalous frame $X^t_A$ within the masked anomalous regions. The term $N$ captures the reconstruction error when the model tries to reconstruct the pseudo-anomalous frame within these same regions. The parameter $\epsilon$ is a small constant to prevent division by zero, thus ensuring numerical stability.

The essence of this loss function is to compel the model to prefer transforming pseudo-anomalous frames back into their normal state. In simpler terms, when the model encounters an anomalous frame, the goal is for its reconstructed output to bear a closer resemblance to a normal frame rather than retaining the anomalous characteristics. Though this is a hopeful outcome for any standard reconstruction model, the distinction loss explicitly trains the model to target this outcome, evidence of this is shown in Supplmentary Material, Section \ref{sec:qualitative} within Figures \ref{fig:ped2_qualatative} and \ref{fig:avenue_qualatative}.

\begin{figure}[t!]
\begin{center}
\includegraphics[scale=.22]{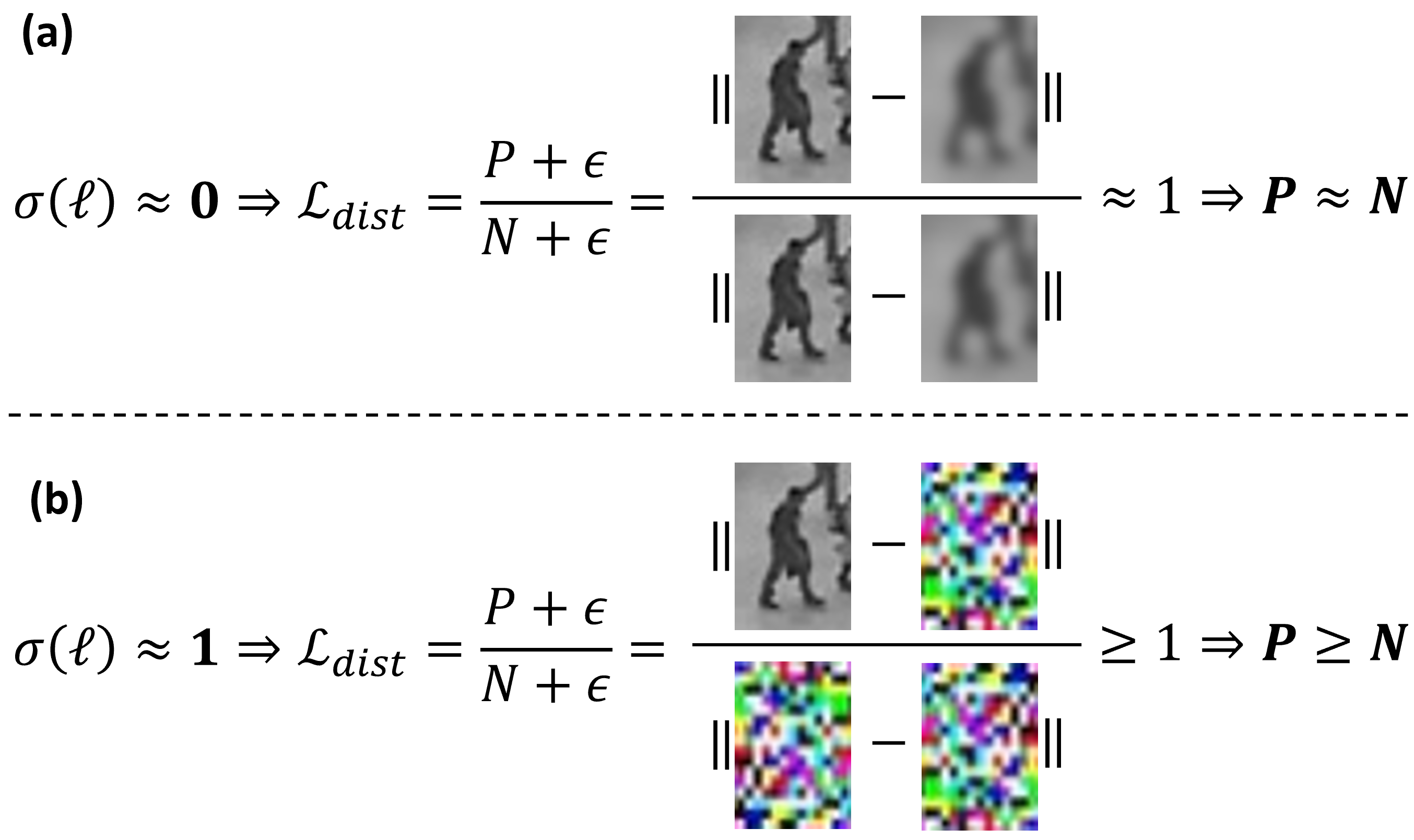}
\caption{Panel (a) depicts a scenario where $\sigma(\ell)$ approaches zero, leading to minimal deviation from the original frame and challenging the model's ability to distinguish between normal and anomalous regions due to the lack of significant noise. Panel (b) illustrates the opposite extreme, where $\sigma(\ell)$ is near one, resulting in an overly distorted anomalous region dominated by noise, which challenges the model's reconstruction capabilities and undermines the distinction loss's effectiveness.}
\label{fig:extremes}
\end{center}
\end{figure}

The underlying intuition of the distinction loss $\mathcal{L}_{dist}$ is to foster a reconstruction process that pulls the pseudo-anomalous frame towards the normal frame more than it does towards itself. This is achieved by aiming to reduce $P$—the difference between the normal frame and its reconstruction from a pseudo-anomalous input—and to increase $N$—the difference between the pseudo-anomalous frame and its reconstruction. By doing so, the model is incentivized to differentiate between normal and anomalous frames, thereby enhancing its anomaly detection capabilities.
This approach contrasts with methodologies employed by our competitors \cite{Astrid2022,Astrid2023}, who focus on maximizing the discrepancy between pseudo anomalous inputs and their reconstructions. 
Such a strategy often results in the emergence of unusual patches within the reconstructed images, a side effect not observed with our model. In contrast, the distinction loss aims to transform pseudo anomalies to resemble normalcy. A visual representation of the distinction loss can be seen in Figure \ref{fig:ddl_architecture}.

For the model's reconstruction function $f$, the ideal scenario is to replicate the normal regions with high fidelity while transforming the anomalous regions towards normalcy. This ability is reflected in the dynamics of $P$ and $N$: 

\begin{itemize}
\item A lower $P$ indicates the model's proficiency in reconstructing the normal aspects of a frame, even when presented with a pseudo-anomalous input.
\item A higher $N$ indicates that the model is not simply replicating the anomalous features present in the pseudo-anomalous frames, but rather is challenged to reconstruct those features, reflecting a discrepancy between the input and the output. 
\end{itemize}

The impact of $\sigma(\ell)$, the anomaly weighting factor, on the distinction loss is pivotal:

\begin{itemize}
\item With $\sigma(\ell)$ approaching zero, the noise's influence on $X_A$ is minimized, leading to a scenario where $X_A$ is almost identical to $X$. This presents a challenge in distinguishing between normal and anomalous frames, as $P$ and $N$ become similar, pushing $\mathcal{L}_{dist} \approx 1$. This can be visualized in Figure \ref{fig:extremes} (a).
\item On the other hand, as $\sigma(\ell)$ tends towards one, the anomalous region is replaced with something which almost entirely resembles noise. The model then faces the nearly impossible challenge of reconstructing the anomalous regions, thus rendering the distinction loss redundant, as shown in Figure \ref{fig:extremes} (b). 
\end{itemize}

However, striking the right balance for $\sigma(\ell)$ is essential: it should be low enough so that the model, $f$, is able to reconstruct normality from a pseudo anomalous frame, but not so low where the pseudo-anomalous frame is too similar to the normal frame; causing $\mathcal{L}_{dist} \approx 1$. The adjustment of $\sigma(\ell)$ is carried out through backpropagation during training, allowing the model to iteratively find the optimal balance to maximize its proficiency in anomaly detection, aiming to pinpoint the smallest discernible anomaly from normalcy.

\subsection{Inference}
During the inference phase, the components involved in training, specifically the anomaly weight, object detection and tracking, and the Pseudo Anomaly Creator, are not utilized. The inference stage is streamlined to function through a conventional reconstruction approach. This process entails imposing a sliding window across each video, then submitting a sequence of video frames directly into the reconstruction model, which then processes these frames to output a reconstructed version of the middle frame. 

\section{Datasets}
\label{sec:datasets}

Our investigation utilizes a suite of video datasets to evaluate the adaptability and effectiveness of our proposed pseudo-anomalous loss approach in more complex scenarios. Specifically, we focus on three prominent video datasets: Ped2, CUHK Avenue, and ShanghaiTech. These datasets, with their varied and intricate anomaly instances, offer a robust testing ground to assess the performance of our model under diverse conditions.

The Ped2 dataset \cite{Lu2013}, sourced from pedestrian area surveillance footage, is notable for its range of anomalous events such as biking, skating, or irregular movement patterns. This dataset provides video clips with a frame resolution of $360\times240$, enabling a diverse sampling environment for anomaly detection research.

The CUHK Avenue dataset \cite{Sabokrou2018}, originating from surveillance systems at the Chinese University of Hong Kong's Avenue, documents typical anomalies like running, loitering, and object throwing. These activities are unusual for the setting, making it an ideal dataset for testing anomaly detection models. Videos in this dataset are presented at a resolution of $640\times360$, offering a detailed view for analysis.

Comprising surveillance footage from a variety of indoor and outdoor scenes, the ShanghaiTech dataset \cite{liu2018ano_pred} introduces a wide range of anomalies, including burglary, climbing, and fighting. The dataset's videos feature a resolution of $856\times480$, with variable frame numbers across clips. This diversity makes the ShanghaiTech dataset a comprehensive platform for challenging and evaluating the capabilities of anomaly detection systems.
\section{Results}
\label{sec:results}


Our experimental setup and performance evaluation, aligning with established benchmarks in anomaly detection, leverages FastRCNN \cite{Girshick_2015_ICCV} for object detection and OCSort \cite{Cao_2023_CVPR} for object tracking during the training phase. Notably, our Conv3DSkipUNet (C3DSU) model processes sequences of 3 frames. We benchmark our model against leading competitors identified in the comprehensive review by Astrid et al. \cite{Astrid2023}, implementing a median window filtering approach with a window size of 17, as effectively demonstrated by Liu et al. \cite{Liu2023}. It is crucial to note that, while our innovative approach leverages object detection to generate pseudo-anomalies during the training phase, the core functionality of our model during inference strictly adheres to the principles of reconstruction-based anomaly detection. Though object detection methods have shown superior performance on the datasets described in Section \ref{sec:datasets}, they are limited in practical application due to their incapability of detecting non-object related anomalies, such as explosions or debris falling off of buildings. Therefore, we strictly compare our methodology to other reconstruction-based methods. This strategic choice differentiates our work from methods reliant on object detection or frame prediction techniques for anomaly identification. 

To quantitatively assess the performance of our model in video anomaly detection tasks, we employ a detailed anomaly scoring mechanism. Each frame's anomaly score is derived by computing the Euclidean distance at the pixel level between the frame and its reconstructed counterpart. To refine this evaluation, the calculated distances are segmented into patches sized $16 \times 16$, with the frame score determined by the highest mean value among these patch scores. 

\begin{table}[ht]
\centering
\begin{tabular}{llccc}
\hline
\textbf{Method} &  \textbf{Year}&\textbf{Ped2} & \textbf{Avenue} & \textbf{SHT} \\
\hline
AE-Conv2D \cite{Hasan2016} &  2016 & 90.00 & 70.20 & 60.85 \\
AE-Conv3D \cite{Zhao_2017} &  2017 & 91.20 & 71.10 & - \\
AE-ConvLSTM \cite{luo2017} &  2017 & 88.10 & 77.00 & - \\
TSC \cite{Luo_2017_iccv} &  2017 & 91.03 & 80.56 & 67.94 \\
StackRNN \cite{Luo_2017_iccv} &  2017 & 92.21 & 81.71 & 68.00 \\
MemAE \cite{Gong2019} &  2019 & 94.10 & 83.30 & 71.20 \\
MNAD \cite{Park2020}&  2020 & 90.20 & 82.80 & 69.80 \\
PseudoBound \cite{Astrid2023} &  2023 & 98.44 & 87.10 & 73.66 \\
MAMC \cite{NING2024159} &  2024 & 96.70 & 87.60 & 71.50 \\
C$^2$Net \cite{Liang2024} &  2024 & 98.00 & 87.50 & - \\
\textbf{C3DSU with DDL} & Ours & \textbf{98.46} & \textbf{90.35} & \textbf{74.25} \\
\hline
\end{tabular}
\caption{Comparative AUC Scores across Ped2, Avenue, and ShanghaiTech datasets. The table presents the AUC performance of our DDL model against a range of competing methodologies, highlighting the best performing results in bold.}
\label{tab:auc_scores}
\end{table}

The performance of our proposed methodology, as quantified through the Area Under the Curve (AUC) scores across three benchmark datasets, demonstrates its superior capability in detecting anomalies within video sequences. Table \ref{tab:auc_scores} showcases a comparative analysis of our model, denoted as C3DSU with DDL, against a variety of established methods in the field.

On the Ped2 dataset \cite{Lu2013}, our approach achieves an AUC score of \textbf{98.46}\%, surpassing the previous state-of-the-art, PseudoBound \cite{Astrid2023}, by a slight margin. This indicates an improvement in the model's ability to detect anomalies, illustrating the effectiveness of the dynamic anomaly weighting and distinction loss mechanism implemented in our methodology.

In the context of the Avenue dataset \cite{Sabokrou2018}, our DDL model demonstrates a notable leap in performance, registering an AUC score of \textbf{90.35}\%. This represents not only an improvement over the PseudoBound method \cite{Astrid2023} but also a substantial advancement compared to other reconstruction-based approaches such as MemAE \cite{Gong2019} and MNAD-Reconstruction \cite{Park2020}. The results underscore our method's adeptness at handling the dataset's complex anomaly scenarios, further establishing the efficacy of incorporating pseudo-anomalies in training to enhance anomaly detection accuracy.


For the ShanghaiTech (SHT) dataset \cite{liu2018ano_pred}, our model achieves an AUC of \textbf{74.25\%}, representing the best performing model amongst those compared. It is important to note that, in addressing the SHT dataset's diverse and dynamic anomaly instances, we trained a unique model for each scene, acknowledging that each scene warrants a different anomaly weight, $\sigma(\ell)$. This scene-specific approach allows for a more tailored anomaly detection mechanism, catering to the unique characteristics and challenges of each scene. The median score of all scenes is then taken to represent the overall performance on the SHT dataset. This methodological nuance underscores the adaptability of our approach, demonstrating its robustness across varied surveillance contexts despite the inherent challenges of the SHT dataset.

\section{Ablation Studies}
\label{sec:ablations}


To elucidate the impact of Dynamic Distinction Learning (DDL) on video anomaly detection, we conducted ablation studies comparing the performance of two models, UNet and Conv3DSkipUNet (C3DSU), on the Ped2 and Avenue datasets, both with and without the implementation of DDL. The UNet model serves as a baseline, employing a traditional architecture without the convolutional 3D (Conv3D) layers between skip connections, and processes single frames independently. In contrast, the C3DSU model, designed for temporal data analysis, incorporates Conv3D layers between skip connections to capture temporal dynamics between frames.

The terminology used to describe the training configurations of the models—specifically, `without DDL', `with SDL', and `with DDL'—reflects the incorporation of our Dynamic Distinction Learning (DDL) framework at different levels. The `without DDL' configuration represents the standard reconstruction training process where the models, UNet and Conv3DSkipUNet (C3DSU), are trained purely on the task of reconstructing normal frames, leveraging only the reconstruction loss and omitting the introduction of pseudo anomalies. In contrast, the `with SDL' (Static Distinction Learning) setup incorporates both the reconstruction loss and a static version of the distinction loss, where the anomaly weight, $\sigma(\ell)$, is fixed at 0.5 and not subject to training adjustments. This static distinction approach aims to introduce a consistent level of challenge in distinguishing anomalies but lacks the adaptability of dynamic weighting. Finally, `with DDL' employs our proposed methodology, integrating the dynamic anomaly weighting mechanism alongside the distinction loss into the training of the models. 

\begin{table}[ht]
\centering
\begin{tabular}{lrlr}
\hline
\textbf{Model} & \textbf{without DDL} & \textbf{with SDL}& \textbf{with DDL}\\
\hline
UNet & 86.90  & 95.28 & \textbf{97.76}\\
C3DSU & 95.55  & 97.12 & \textbf{98.46}\\
\hline
\end{tabular}
\caption{This table illustrates the performance improvement on the Ped2 dataset facilitated by the Dynamic Distinction Learning (DDL) approach across two different architectures: UNet and C3DSU.}
\label{tab:ped2_performance_comparison}
\end{table}

As shown in Table \ref{tab:ped2_performance_comparison}, the implementation of DDL significantly enhances model performance. For the UNet model, the Area Under the Curve (AUC) score increases from 86.90\% without DDL to 95.28\% with SDL and further to \textbf{97.76\%} with DDL, underscoring the effectiveness of DDL in enhancing anomaly detection accuracy. The introduction of a static distinction loss already marks a notable improvement, demonstrating the value of integrating anomaly differentiation into the training process. Similarly, the C3DSU model benefits from the addition of DDL, with its AUC score improving from 95.55\% to 97.12\% with SDL and then to \textbf{98.46\%}. These results highlight the pivotal role of DDL in refining the model's ability to differentiate between normal and anomalous frames, particularly when temporal dynamics are considered. The improvement seen with SDL indicates the initial benefits of incorporating distinction mechanisms, which are significantly amplified upon transitioning to dynamic weighting.

The impact of DDL is also evident in the performance on the Avenue dataset, as depicted in Table \ref{tab:avenue_performance_comparison}. The UNet model experiences an improvement in AUC score from 84.18\% without DDL to 87.06\% with SDL and further to \textbf{88.96\%} with DDL. The introduction of SDL showcases a tangible improvement, setting the stage for the more substantial enhancements afforded by the dynamic approach. The C3DSU model, however, showcases a more pronounced improvement, with the AUC score increasing from 82.54\% without DDL to 89.41\% with SDL and then to \textbf{90.35\%} with DDL. These findings demonstrate the utility of DDL across different architectural frameworks and datasets, especially in scenarios involving complex anomaly patterns. The step-wise enhancements from static to dynamic distinction learning illustrate the methodological progression and its impact on the models' anomaly detection capabilities, highlighting the critical role of adaptively learning the anomaly weight for maximizing detection accuracy.

\begin{table}[ht]
\centering
\begin{tabular}{lrlr}
\hline
\textbf{Model} & \textbf{without DDL} &\textbf{with SDL}& \textbf{with DDL}\\
\hline
UNet & 84.18  & 87.06 & \textbf{88.96}\\
C3DSU & 82.54  & 89.41 & \textbf{90.35}\\
\hline
\end{tabular}
\caption{This table presents a comparison of model performance on the Avenue dataset, with and without the incorporation of Dynamic Distinction Learning (DDL), across UNet and C3DSU architectures.}
\label{tab:avenue_performance_comparison}
\end{table}


The ablation studies highlight the incremental value offered by each component of our methodology. The introduction of the distinction loss with a static pseudo anomaly weight significantly improves model performance by explicitly training the model to map pseudo anomalies towards normality. Further refinement is achieved with the implementation of a dynamic anomaly weight, which empowers our methodology to adaptively fine-tune and identify the minimum level of anomaly that can be detected. 
\section{Conclusion}

This paper introduced Dynamic Distinction Learning (DDL), a novel approach designed to enhance the accuracy of video anomaly detection through the integration of pseudo-anomalies, dynamic anomaly weighting, and a unique distinction loss function. Our methodological innovation lies in its ability to adaptively learn the variability of normal and anomalous behaviors without relying on fixed anomaly thresholds, thereby significantly improving detection performance.

Our experiments, conducted on benchmark datasets such as Ped2, CUHK Avenue, and ShanghaiTech, have demonstrated the superior performance of the DDL framework. The model achieved remarkable AUC scores, outperforming existing state-of-the-art methods on the Ped2 and Avenue datasets, and delivering competitive results on the ShanghaiTech dataset. These achievements underscore the effectiveness of DDL in addressing the challenges of video anomaly detection, offering a scalable and adaptable solution that can be tailored to specific scene requirements.

The ablation studies further highlighted the impact of incorporating DDL into different model architectures, including UNet and Conv3DSkipUNet (C3DSU). The significant improvements in anomaly detection accuracy with DDL underscore its role in refining models' ability to distinguish between normal and anomalous events effectively, showcasing its broad applicability across different architectural frameworks and complex anomaly patterns.

In conclusion, Dynamic Distinction Learning represents a significant advancement in the field of video anomaly detection. Its ability to dynamically adapt and learn from pseudo-anomalies, coupled with the distinction loss function, provides a robust framework for accurately identifying anomalous events in video data.

{
    \small
    \bibliographystyle{ieeenat_fullname}
    \bibliography{main}
}

\clearpage
\setcounter{page}{1}
\maketitlesupplementary

\begin{figure*}[t!]
\begin{center}
\includegraphics[width=\textwidth]{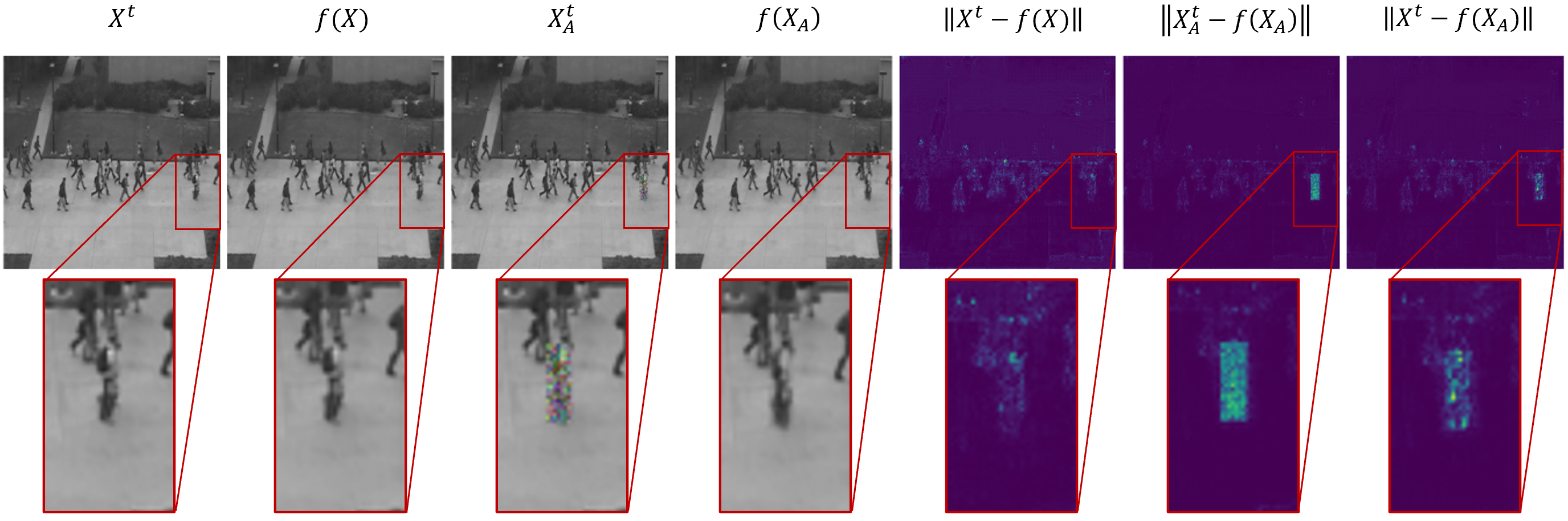}
\caption{Visual Comparison of Model Training Effects: This figure provides a comprehensive visualization of the model's performance across different frames and stages of reconstruction. It features the original middle frame $X^t$, the reconstructed frame from normal input $f(Xt$, the pseudo-anomalous middle frame $X_A^t$, the reconstructed frame from the pseudo-anomalous input $f(X_A)$, and the reconstruction errors $\lVert X^t-f(X) \rVert$, $\lVert X_A^t-f(X_A) \rVert$, and $\lVert X^t-f(X_A) \rVert$.}
\label{fig:training_examples}
\end{center}
\end{figure*}

\subsection{Dynamics of Anomaly Weight $\sigma(\ell)$ in Model Training (Methodology Supplementary)}

To delve deeper into the dynamics of our model's training, it is imperative to scrutinize the behavior of the anomaly weight $\sigma(\ell)$, which serves as a crucial variable in the generation of pseudo-anomalous frames. This parameter, $\sigma(\ell)$, is not merely a static coefficient but a dynamic element that evolves during the training process, reflecting the model's progressing ability to discern between normal and anomalous instances.

\begin{figure}[b!]
\begin{center}
\includegraphics[scale=.23]{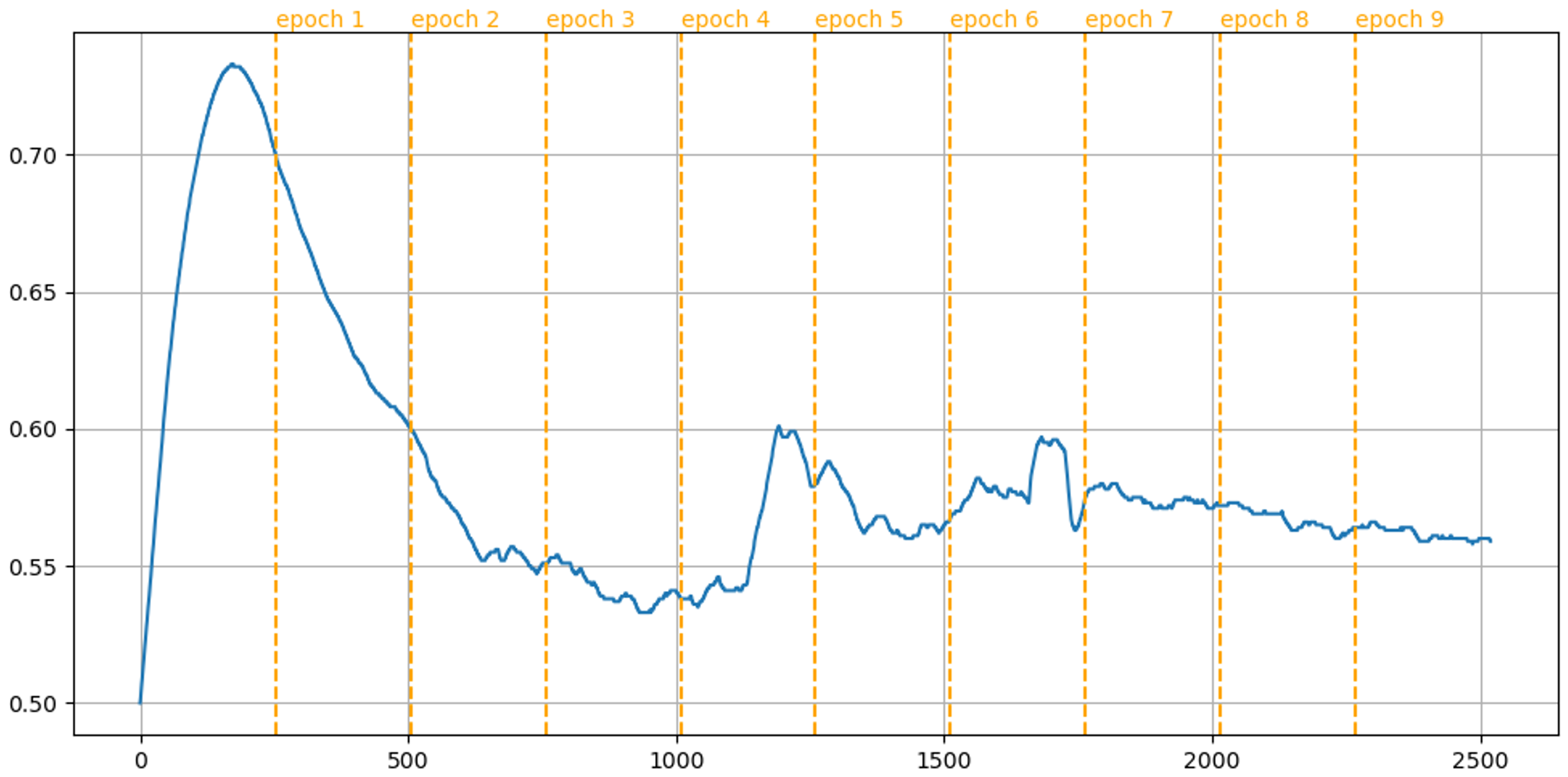}
\caption{Evolution of the anomaly weight $\sigma(\ell)$ during the training of the C3DSU model on the Ped2 Dataset for 10 epochs.}
\label{fig:anomaly_weight_change}
\end{center}
\end{figure}

As training progresses, the trend in the anomaly weight $\sigma(\ell)$ reflects the model's adaptation and learning curve. Initially, $\sigma(\ell)$ may experience fluctuations, including a rapid increase as the model begins to differentiate between normal and anomalous patterns - as it has in Figure \ref{fig:anomaly_weight_change} when training the C3DSU model on the Ped2 Dataset. This early rise in $\sigma(\ell)$ is crucial as it indicates the model's initial phase of learning to handle more pronounced anomalies. As the model's ability to reconstruct and identify anomalies improves, we observe a subsequent decline and eventual stabilization of $\sigma(\ell)$, suggesting that the model is settling on an optimal threshold for anomaly detection without being overly influenced by the injected noise levels.

The crux of training with $\sigma(\ell)$ lies in pinpointing the minimal anomaly magnitude that still allows our model $f$ to reliably differentiate anomalous from normal. This threshold is the 'sweet spot' where $\sigma(\ell)$ is neither too insubstantial to be deemed noise nor too dominant to mask the underlying structure of the data. It's at this juncture that our model $f$ is optimally trained to flag deviations from normalcy, while remaining anchored enough to not be swayed by random perturbations or outliers.

In essence, the evolution of $\sigma(\ell)$ during the training is a barometer of the model's growing intelligence in anomaly detection. By carefully calibrating this parameter, we empower our model $f$ to identify the minimum anomaly required to discern abnormalities, a testament to its analytical prowess and the culmination of a successful training regimen.

\subsubsection{Visualizing the Effects of Training}

To further illustrate the success of our training process and the dynamic adjustment of the anomaly weight $\sigma(\ell)$, Figure \ref{fig:training_examples} presents a visual comparison. These images encapsulate various aspects of our model's performance, including the original middle frame $X^t$, the reconstruction from the normal input $f(X)$, the pseudo-anomalous middle frame $X_A^t$, the reconstruction from the pseudo-anomalous input $f(X_A)$, and the respective reconstruction errors $\lVert X^t-f(X) \rVert$, $\lVert X_A^t-f(X_A) \rVert$, and $\lVert X^t-f(X_A) \rVert$.

This visual representation demonstrates the impact of our training methodology. Particularly, the comparison between $\lVert X_A^t-f(X_A) \rVert$ and $\lVert X^t-f(X_A) \rVert$ reveals a critical insight: the discrepancy between the reconstruction of the pseudo-anomalous frame and the normal frame is significantly less than that between the reconstruction of the pseudo-anomalous frame and its corresponding anomalous input. This outcome underscores the model's capability to more closely align the reconstructed output with the normal frame rather than perpetuating the anomalies present in the pseudo-anomalous input.

\subsection{Rationale Behind Weighted Noise for Simulating Anomalies (Methodology Supplementary)}
\label{subsec:rationale_weighted_noise}

The efficacy of employing weighted noise to simulate a range of human-defined anomalies -- such as skipped frames, duplicate frames, random patches, and the insertion of foreign shapes or objects -- is rooted in the fundamental operating principles of convolutional neural networks (CNNs) employed in reconstruction-based anomaly detection methods. These networks are adept at learning and reconstructing patterns observed in the training data, which predominantly consists of normal behavioral patterns within video sequences. 

When confronted with anomalies, the convolutional kernels of a CNN do not perceive these as distinct types of irregularities per se, but rather as inputs lacking the regular patterns or structures they have been trained to recognize and reconstruct. From the perspective of these kernels, anomalies disrupt the spatial and temporal consistency of the input data, rendering them as pattern-less examples -- essentially, noise. This perception is crucial for understanding why weighted noise can serve as a universal proxy for various anomalies in training anomaly detection models.

The concept of weighted noise as a universal anomaly proxy is further justified by the intrinsic adaptability and learning mechanisms of CNNs. These networks, through their deep architecture, are designed to capture and encode complex patterns in the data they process. The introduction of weighted noise challenges these networks in a unique way, compelling them to discern between the `normal' patterns they've learned to reconstruct and the `abnormal' patterns represented by the noise. This challenge is akin to exposing the network to a wide variety of anomalies without the need for explicit enumeration or replication of each possible anomalous event, which in real-world applications is infeasible due to the vast and unpredictable nature of such events.

\subsection{C3DSU Architecture (Results Supplementary)}
\label{C3DSU_Arch}

The design of our Conv3DSkipUNet (C3DSU) model incorporates a Conv2D UNet structure, enhanced with custom ConvBlocks for both the Encoder and Decoder components, and augmented by the integration of Conv3D layers within the skip connections for handling the temporal dimension. 

\begin{figure}[h!]
\begin{center}
\includegraphics[scale=.23]{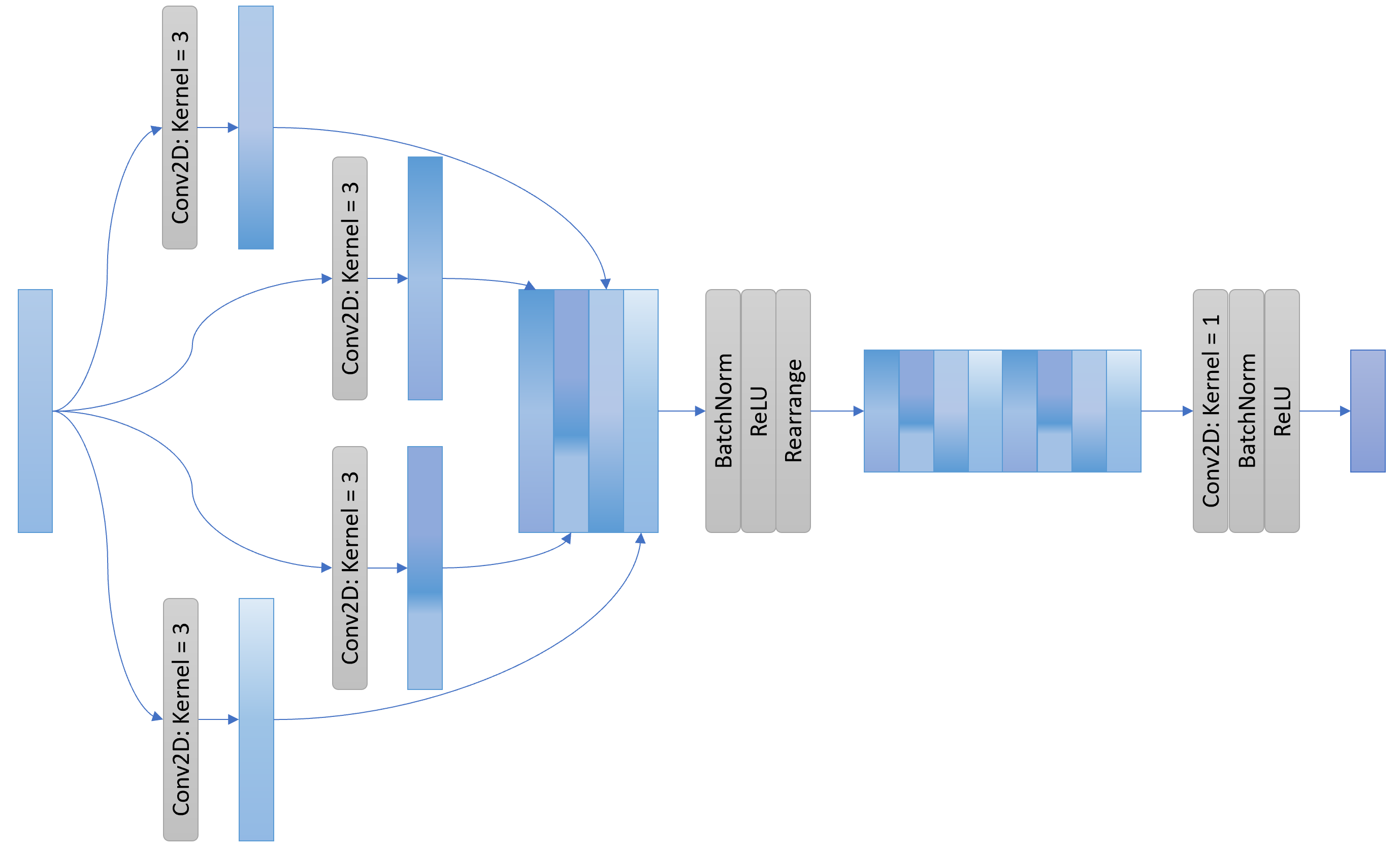}
\caption{Schematic illustration of an Encoder ConvBlock within the C3DSU architecture, highlighting the multi-headed convolution process. This diagram details the structure and flow through the ConvBlock, including the initial parallel Conv2D layers, concatenation, batch normalisation, ReLU activation, dimension reduction, and final Conv2D layer, followed by another round of batch normalisation and ReLU activation.}
\label{fig:EncConvBlock2D}
\end{center}
\end{figure}

A ConvBlock in this context is engineered to facilitate multi-headed convolution. Specifically, it comprises four Conv2D layers, each employing "same" padding to maintain dimensional consistency and a kernel size of 3 for capturing spatial details. These layers are executed in parallel and their outputs are concatenated along the channel dimension, ensuring a rich feature representation. Following the concatenation, the process involves batch normalisation and activation through the ReLU function, aiming to stabilize learning and introduce non-linearity, respectively. The output is then restructured to reduce the Height and Width dimensions, the output is then passed through an additional Conv2D layer with "same" padding, followed by another round of batch normalisation and ReLU activation. The Encoder ConvBlock's operational flow is depicted in Figure \ref{fig:EncConvBlock2D} for visual clarification.

\begin{figure}[h!]
\begin{center}
\includegraphics[scale=.28]{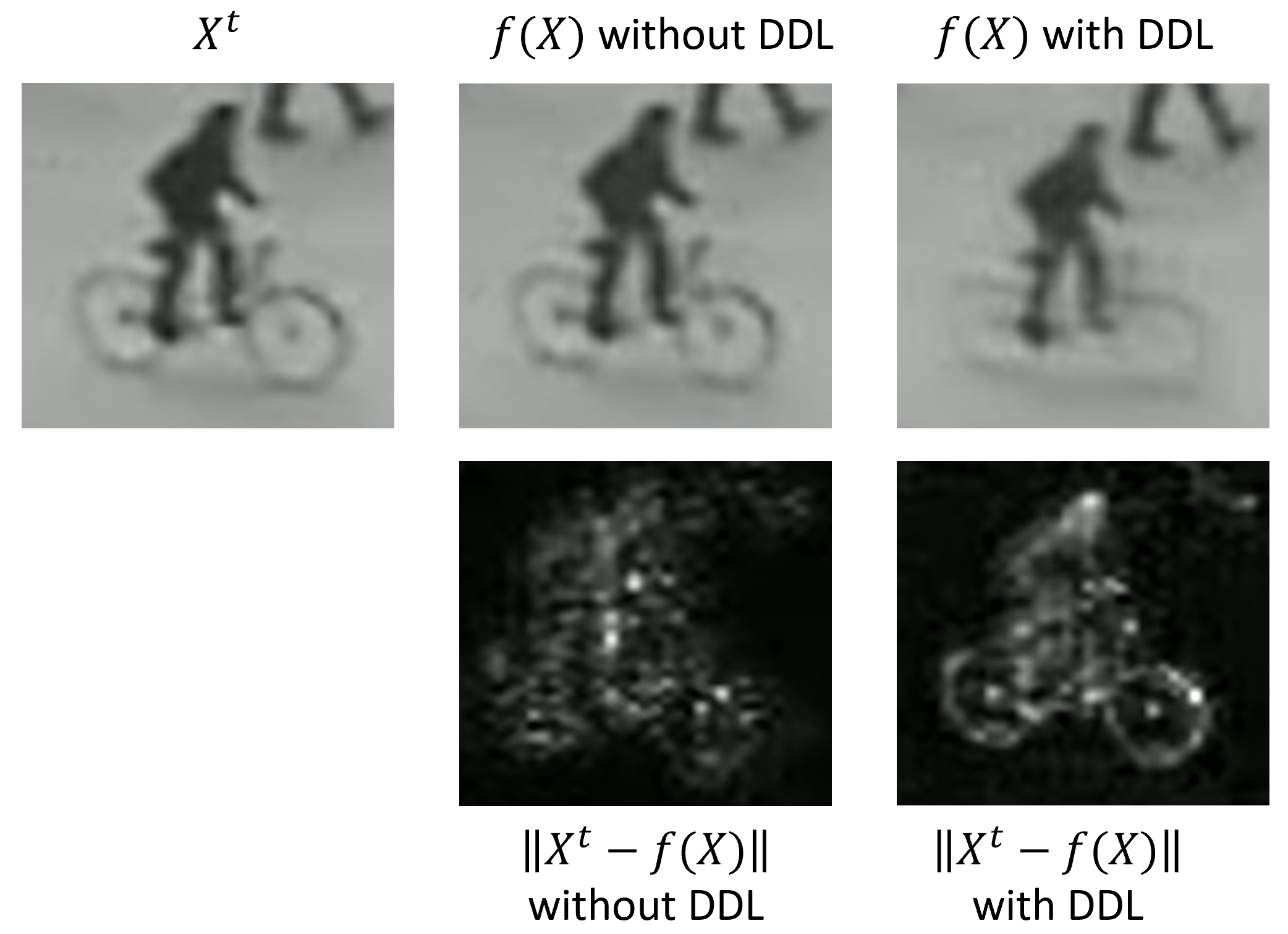}
\caption{Visual anomaly detection comparison in the Ped2 dataset featuring a cyclist riding a bicycle. From left to right: the original frame, reconstruction without Dynamic Distinction Learning (DDL), reconstruction with DDL, and beneath the residual difference highlighting the anomaly.}
\label{fig:ped2_qualatative}
\end{center}
\end{figure}

The overarching UNet architecture is assembled with eight such ConvBlocks, evenly split between the Encoder and Decoder. Each ConvBlock in the Encoder is paired with a corresponding ConvBlock in the Decoder via skip connections. These connections are uniquely designed to pass through Conv3D layers, thereby incorporating the temporal aspect into the spatial information flow. This mechanism ensures that while individual frames are initially processed as separate images by the Conv2D layers in the Encoder and Decoder, the Conv3D layers within the skip connections facilitate the integration of temporal information, essential for effective video analysis.

\subsection{Qualitative Results (Ablations Supplementary)}
\label{sec:qualitative}

\begin{figure}[h!]
\begin{center}
\includegraphics[scale=.28]{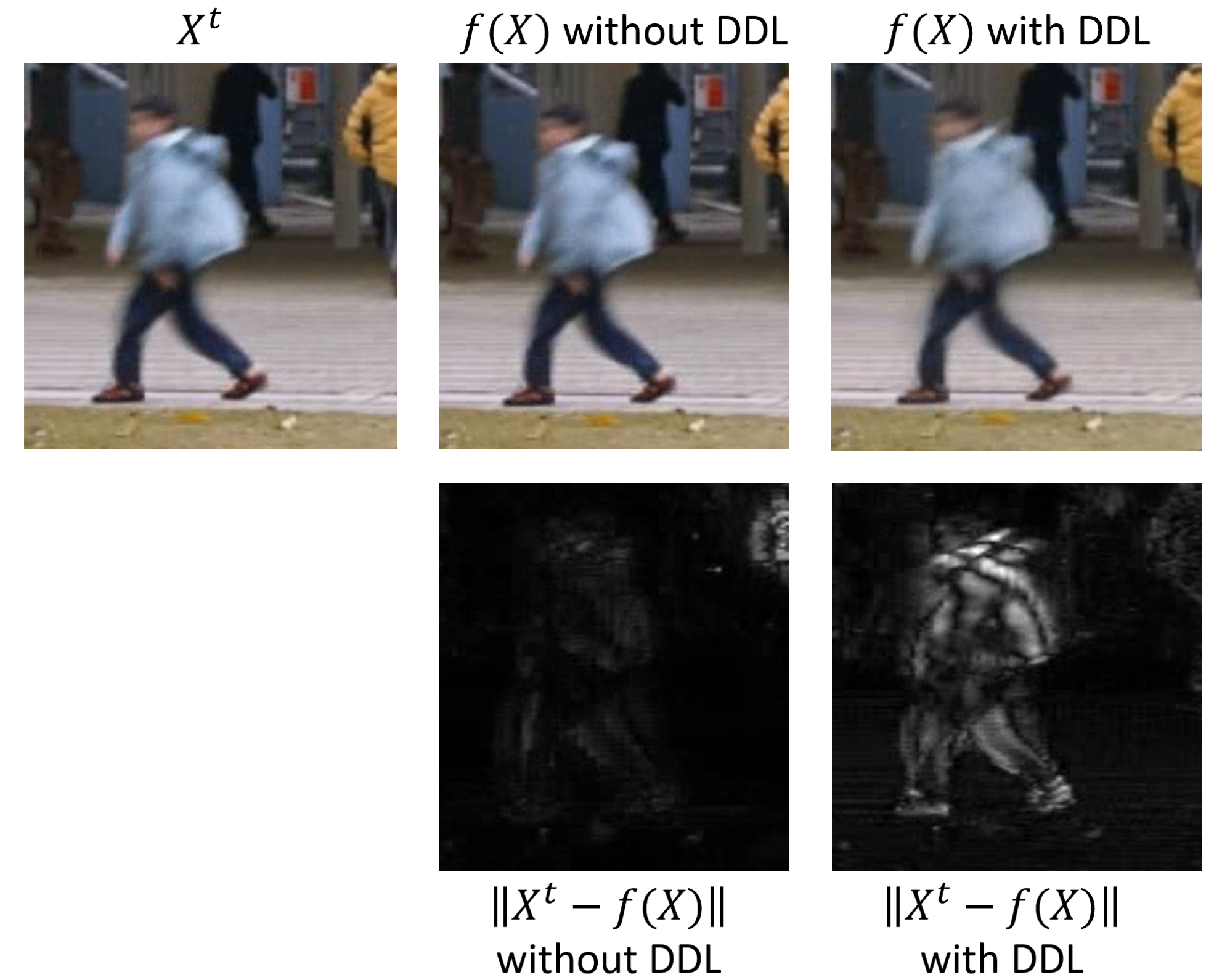}
\caption{Visual anomaly detection comparison in the Avenue dataset illustrating a boy skipping. From left to right: the original frame, reconstruction without Dynamic Distinction Learning (DDL), reconstruction with DDL, and beneath the residual difference emphasizing the anomaly.}
\label{fig:avenue_qualatative}
\end{center}
\end{figure}

To complement the quantitative analysis, qualitative assessments were conducted to visually inspect the model's performance in identifying anomalies within the Ped2 and Avenue datasets. These assessments provide insight into the model's ability to reconstruct scenes and highlight anomalous activities when DDL is applied versus when it is not.

Figure \ref{fig:ped2_qualatative} showcases a visual comparison of anomaly detection in a scenario involving a cyclist riding a bicycle, an anomalous event within the Ped2 dataset. The sequence displays the original frame, followed by reconstructions without and with DDL, and finally, the residual differences between the reconstructions and the original frame. Notably, the application of DDL results in a reconstruction where the bicycle is almost entirely erased, signifying the model's training to poorly reconstruct unfamiliar shapes. This illustrates DDL's effectiveness in forcing the model to focus on normal patterns, thereby making anomalies, such as the bicycle in this case, more pronounced.

Similarly, Figure \ref{fig:avenue_qualatative} presents a visual analysis involving a boy skipping, an anomalous event in the Avenue dataset. The illustration includes the original frame along with reconstructions without and with DDL, supplemented by the residual differences highlighting the anomaly. The comparison clearly demonstrates that with DDL, the anomaly of the boy skipping is accentuated more effectively than without DDL. This enhancement is evident in the residual images, where DDL's reconstruction struggles more with the skipping motion, thereby amplifying the distinction from normal activity.


\end{document}